\documentclass[10pt,twocolumn,letterpaper,table,xcdraw]{article}

\usepackage{cvpr}              
\usepackage[dvipsnames]{xcolor}

\usepackage[accsupp]{axessibility}
\usepackage{multirow}
\usepackage[misc]{ifsym}
\definecolor{cvprblue}{rgb}{0.21,0.49,0.74}
\usepackage[pagebackref,breaklinks,colorlinks,citecolor=cvprblue]{hyperref}
\usepackage{pifont}
\usepackage{tcolorbox}
\usepackage{listings}

\usepackage{amsmath}

\DeclareMathOperator*{\argmin}{arg\,min}

\def\paperID{9445}
\def\confName{CVPR}
\def\confYear{2024}

\title{Weakly-Supervised Emotion Transition Learning for \\ Diverse 3D Co-speech Gesture Generation}

\author{Xingqun Qi$^{1}$, Jiahao Pan$^{1}$, Peng Li$^{1}$, Ruibin Yuan$^{1}$, Xiaowei Chi$^{1}$, Mengfei Li$^{1}$\\
Wenhan Luo$^{1}$, Wei Xue$^{1}$, Shanghang Zhang$^{2}$, Qifeng Liu$^{1, \textrm{\Letter}}$, Yike Guo$^{1, \textrm{\Letter}}$\\
$^{1}$ The Hong Kong University of Science and Technology \\ 
$^{2}$ Peking University\\
 {\tt\small xingqun.qi@connect.ust.hk, \{liuqifeng, yikeguo\}@ust.hk } \\
}

\begin{document}
\twocolumn[{%
\renewcommand\twocolumn[1][]{#1}%
\maketitle
\vspace{-10mm}
\begin{center}
    \centering
    \includegraphics[width=0.94\linewidth]{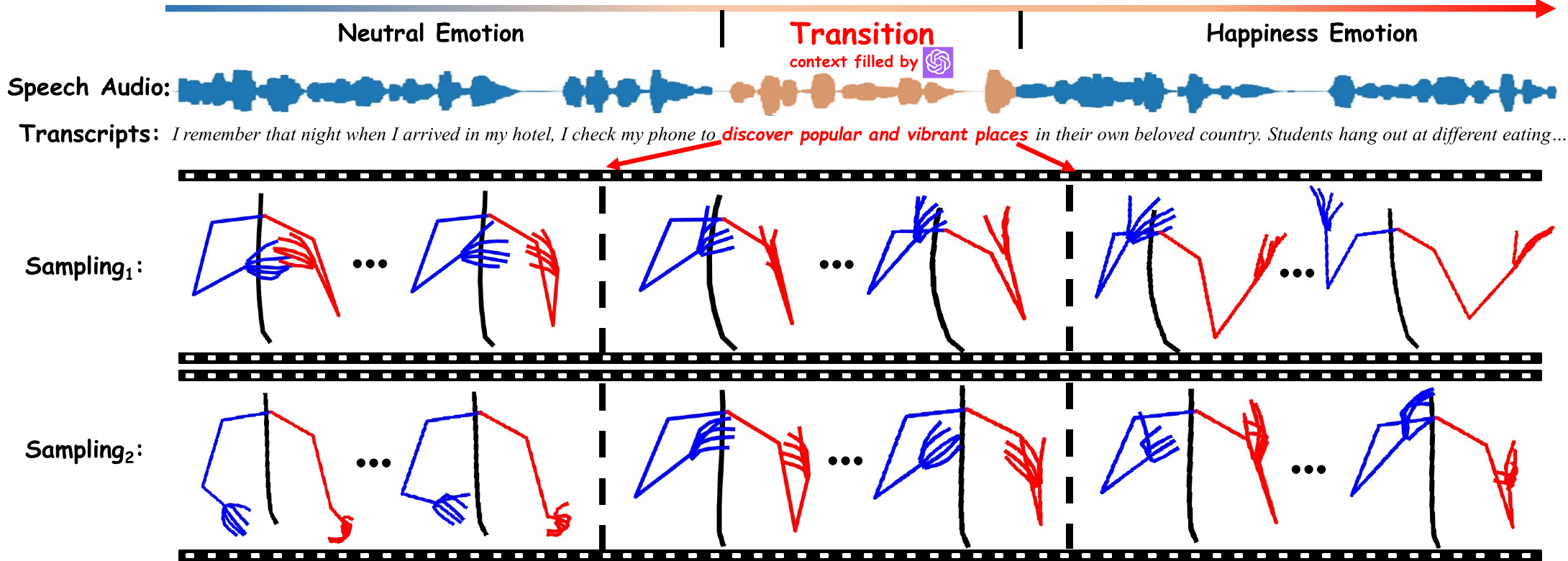}
    \captionof{figure}{Diverse exemplary clips sampled by our method from \textbf{our newly collected BEAT Emotion Transition Dataset}. The vital frames are visualized to demonstrate that the upper body gestures change with the emotion transition of human speech, synchronously. From top to bottom: the input speech audio, the corresponding transcript, and two sampled clips. Best view on screen. }
    \label{fig:teaser}
\end{center}
}]
\renewcommand{\thefootnote}{\fnsymbol{footnote}}
\footnotetext{$\textrm{\Letter}$ Corresponding authors.}

\begin{abstract}
Generating vivid and emotional 3D co-speech gestures is crucial for virtual avatar animation in human-machine interaction applications. While the existing methods enable generating the gestures to follow a single emotion label, they overlook that long gesture sequence modeling with emotion transition is more practical in real scenes. In addition, the lack of large-scale available datasets with emotional transition speech and corresponding 3D human gestures also limits the addressing of this task. To fulfill this goal, we first incorporate the ChatGPT-4 and an audio inpainting approach to construct the high-fidelity emotion transition human speeches. Considering obtaining the realistic 3D pose annotations corresponding to the dynamically inpainted emotion transition audio is extremely difficult, we propose a novel weakly supervised training strategy to encourage authority gesture transitions. 
Specifically, to enhance the coordination of transition gestures \wrt different emotional ones, we model the temporal association representation between two different emotional gesture sequences as style guidance and infuse it into the transition generation. We further devise an emotion mixture mechanism that provides weak supervision based on a learnable mixed emotion label for transition gestures. Last, we present a keyframe sampler to supply effective initial posture cues in long sequences, enabling us to generate diverse gestures. Extensive experiments demonstrate that our method outperforms the state-of-the-art models constructed by adapting single emotion-conditioned counterparts on our newly defined emotion transition task and datasets. Our code and dataset will be released on the project page: \href{https://xingqunqi-lab.github.io/Emo-Transition-Gesture/}{\textit{https://xingqunqi-lab.github.io/Emo-Transition-Gesture/}}.

\end{abstract}    
\section{Introduction}
\label{sec:intro}

Co-speech gesture generation aims to synthesize vivid and emotional human postures coordinated with the audio input. These non-verbal behaviors serve as a key factor during human conversations that significantly facilitates the delivery of speech content~\cite{cassell1999speech,goldin2013gesture,hostetter2008visible}. Meanwhile, modeling co-speech gestures has a wide range of embodied AI applications in human-machine interaction~\cite{koppula2013anticipating,liu2023audio}, robot assistants~\cite{farouk2022studying}, and virtual/augmented reality (AR/VR)~\cite{fu2022systematic}. Conventionally, many researchers usually focus on synthesizing human upper body gestures consistent with speech audio~\cite{liu2022learning, zhu2023taming, yoon2020speech, yi2023generating, yang2023diffusestylegesture}.

Nevertheless, except for a few recent works that generate the co-speech gestures of a single emotion category~\cite{liu2022beat,zhi2023livelyspeaker, ao2023gesturediffuclip, bhattacharya2021speech2affectivegestures}, previous works mostly focus on emotion-agnostic generation~\cite{yoon2020speech, yi2023generating, bhattacharya2021text2gestures}. Most of them overlook synthesizing the long sequence co-speech gestures with the emotion transitions, which are more practical in real-world scenes. For example, a person may 
not maintain a single emotion forever when communicating with others or in speech talking. In this work, we therefore introduce the task of \textbf{\emph{speech-driven emotion transition}} for generating vivid and diverse 3D co-speech gestures, displayed in Figure \ref{fig:teaser}. 
There are two main challenges in this task: 1) Datasets of 3D co-speech gestures synchronized with emotion transition speech audios are scarce. Creating such a dataset containing 3D human postures is difficult due to the lack of guidance from emotional experts and complex motion capture systems. 2) Modeling the plausible and temporal coherent co-speech gestures from one emotion to another in long sequences is difficult, especially in the transition duration.

To overcome the issue of data scarcity, we newly present two datasets containing emotion-transition human speech that are built upon previous single-emotion ones~\cite{liu2022learning, liu2022beat}. In particular, thanks to the developed language model ChatGPT-4~\cite{openai2023gpt4}, we first leverage it to generate text transcripts of the transition based on speech context. Then, by employing the audioLDM2 technique~\cite{liu2023audioldm, liu2023audioldm2} for audio inpainting, we ensure the inpainted transition's timbral consistency with its adjacent contexts and a smooth emotional transition throughout. 
Afterwards, to support our insight on modeling co-speech gestures coherent with emotion-transition speech, corresponding 3D human postures of transition are required.

However, due to the dynamically generated transition transcripts, it is infeasible to construct the aligned realistic 3D pose annotation of human bodies.
Hence, we solve the challenges of vivid co-speech gesture generation in a novel weakly supervised pattern, containing a motion transition infusion mechanism and an emotion mixture strategy. Specifically, in the motion transition infusion mechanism, we model the temporal correlation between the generated head and tail gesture features as style guidance representation. The style guidance representation provides motion transition cues that are infused into the transition embedding via an adaptive instance normalization (AdaIN) layer~\cite{huang2017arbitrary}. Along with this manner, we can effectively enhance the coordination of transition gestures \wrt two different emotional ones.

Moreover, to alleviate the lack of supervision during the transition between two emotions, the emotion mixture strategy is built to provide weak emotional supervision of the generated transition gestures. Concretely, we learn a joint embedding of two different emotional gesture sequences using a temporal aggregation encoder. Then, we pre-train an emotion classifier based on the annotated human 3D poses with single emotion labels in the dataset. Here, this joint embedding is leveraged as an emotion mixture weight for the pre-trained classifier to facilitate high-fidelity transition gesture synthesis with desirable properties. Finally, considering the generated 3D postures should be non-deterministic given the human speech, we devise a keyframe sampler to produce diverse initial poses as reference. In this fashion, our method enables diverse co-speech gesture generation with emotion transitions.  Extensive experiments conducted on our newly constructed two datasets verify the effectiveness of our methods, displaying vivid and emotional 3D co-speech gestures.

\noindent Overall, our contributions are summarized as follows:
\begin{itemize}[leftmargin=*]
    \vspace{0.5em}
    \item We introduce a new task of emotion transition co-speech gesture generation cooperating with two newly constructed datasets named BEAT Emotion Transition (BEAT-ETrans) and TED Emotion Transition (TED-ETrans), significantly facilitating research on 3D human motion modeling.
    \vspace{0.5em}
    
    \item 
    We design a motion transition infusion mechanism to ensure the temporal coordination of transition gestures \wrt two different emotional ones and a weakly supervised emotion mixture strategy to enable high-fidelity transition gesture synthesis with desirable properties.
    
    \vspace{0.5em}
    \item Extensive experiments show that our method outperforms state-of-the-art counterparts on both datasets, displaying realistic and vivid co-speech gestures given emotion transition human audios.

\end{itemize}
\section{Related Work}
\label{sec:related}

\subsection{Co-speech Gesture Synthesis}
Synthesizing human co-speech gestures plays a significant role in various applications~\cite{koppula2013anticipating, huang2022proxemics, wang2022self, qi2023diverse}. Numerous studies have been proposed to address these issues that are roughly divided into rule-based approaches~\cite{kipp2005gesture, kopp2006towards}, machine-learning-based approaches~\cite{levine2010gesture, sargin2008analysis}, and deep learning-based ones~\cite{liu2022learning, zhu2023taming, yoon2020speech, yi2023generating,liu2022beat,qi2023emotiongesture, zhi2023livelyspeaker, ao2023gesturediffuclip}. Traditional researchers follow the rule-based workflow, leveraging the speech-gesture pairs as guidance to generate co-speech gestures pre-defined by linguistic experts. Other early works integrate the manually defined gesture features with machine learning techniques to synthesize the co-speech gestures. In the aforementioned two manners, the researchers usually focus on the optimization of the matching process between human speech and pre-defined gestures. It may require experts much expensive effort in the speech-gesture pair construction.

Recently works focus on building the mapping directly from the input human speech and sequential gestures by exploiting the deep neural networks. They usually leverage multi-modality cues to facilitate the generation of co-speech gestures, associating with the speech audio~\cite{Li_2021_ICCV, zhu2023taming, yi2023generating}, speaker identity~\cite{yoon2020speech}, emotion~\cite{liu2022beat, ao2023gesturediffuclip, liu2023emage}, and text transcript~\cite{liu2022learning}. However, they overlook that emotion transition of the long sequential co-speech gesture modeling is much more practical in the real scenes. Moreover, since the lack of annotated 3D gestures corresponding to dynamically constructed transition speech, few of the above methods could be directly adapted to this new thought.

\subsection{3D Human Motion Modeling}
Human motion modeling aims to generate realistic and smooth human motions with various multi-modality conditions~\cite{jiang2023motiongpt, mao2022weakly, endo2023motion}, including co-speech gesture generation as a sub-task. One of the hottest topics is synthesizing human motion from text prompts with a few past postures as the seed~\cite{petrovich2021action, zhao2023diffugesture, voas2023best}. These methods usually engage in forcing human motion to represent the semantic expression aligned with the text. Literally, the task most closely related to ours is AI choreographer~\cite{li2020learning, li2021ai, siyao2022bailando} which generates the motion from music signals. However, the AI choreographer works mainly focus on the rhythmic-coherent motion of the whole human body but without subtle finger gestures. 
While sharing a similar goal with the approaches mentioned above, our newly defined work differs from them significantly. We take the emotion-transition long sequence co-speech gesture without corresponding 3D human pose annotation into consideration, thus motivating us to utilize the motion transition of two annotated emotional gestures and coherent the overall sequence.

\begin{figure*}[t]
\begin{center}
\includegraphics[width=0.92\linewidth]{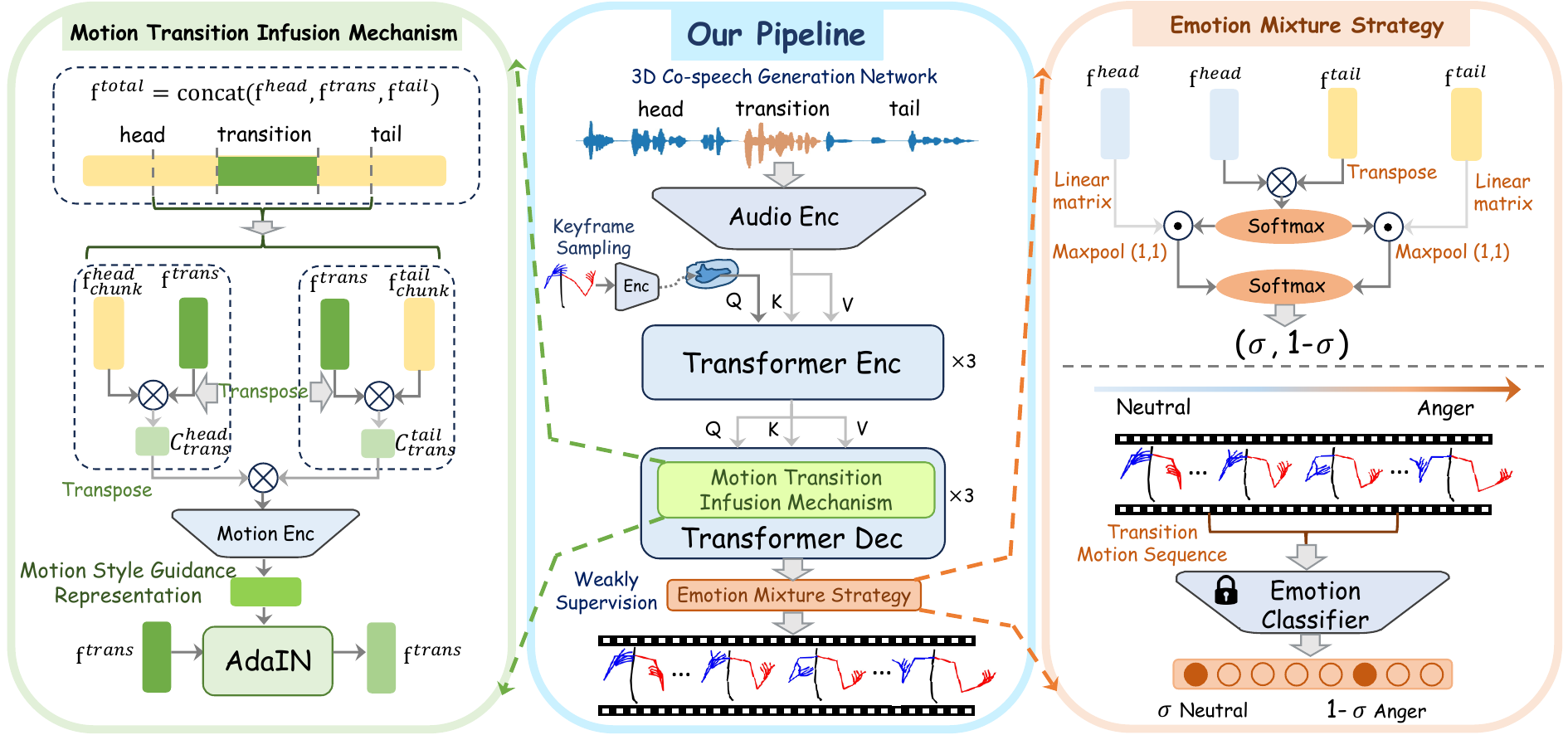}
\end{center}
\vspace{-1em}
\caption{\textbf{The overview of our proposed method.} The \textbf{middle part (\textcolor{blue}{blue})} displays the overall pipeline for 3D co-speech gesture generation from emotion transition human speech. The \textbf{left part (\textcolor{green}{green})} depicts our proposed Motion Transition Infusion Mechanism that enhances the coordination of transition gestures \wrt head/tail ones. The \textbf{right part (\textcolor{orange}{orange})} shows the designed Emotion Mixture Strategy to provide weak supervision of the generated transition gestures, thereby achieving authority producing.
}
\label{fig:pipeline}
\vspace{-0.5em}
\end{figure*}
\section{Proposed Method}
\label{sec:methods}

\subsection{Emotion Transition Dataset Construction}
We aim to address the emotion transition co-speech gesture generation in a weakly supervised manner. Due to the existing paired speech-gesture datasets~\cite{liu2022beat, qi2023emotiongesture}, we could focus on synthesizing the high-fidelity transition human audios. 
Synthesizing datasets conducive to our task focuses on ensuring semantic coherence, smooth emotional audio transitions, consistent timbre, and audio fidelity.

\vspace{0.5em}
\noindent \textbf{Preliminary:}
Considering our key insight to modeling the co-speech gesture with emotion transition, we first split the existing aligned speech-gesture pairs \cite{liu2022beat, qi2023emotiongesture} into four-second clips. To diversify the datasets, we randomly splice two clips from the same speaker to construct an emotion-transition candidate pair. The clip with neutral emotion is leveraged as head speech, and the other with various emotions is represented as tail speech. We leverage the dynamically synthesized two-second audio as a transition to combine the head and tail speeches.
In the following, we briefly summarize our efforts in constructing the dynamic transition speech audio.

\vspace{0.5em}
\noindent \textbf{Textual Inpainting of Transition:} To ensure the semantic coherence of transition \wrt head/tail speeches, we exploit the advanced language generation model ChatGPT-4~\cite{openai2023gpt4} to complete the transcript according to the context. Literally, we follow the conventional estimation that people usually talk 30 phonemes~\cite{roach1998some} in a two-second speech as the prompt to guide transcript generation by ChatGPT-4.

\vspace{0.5em}
\noindent \textbf{Synthesis of Transition Audio:} Once we obtain the transcript of transitions, we employ a superior text-to-speech model, audioLDM2~\cite{liu2023audioldm, liu2023audioldm2}, to generate corresponding speech audio. Here, we leverage the speaker embeddings extracted via SpeechBrain's ECAPA-TDNN \cite{ravanelli2021speechbrain, desplanques2020ecapa} as prior guidance to maintain the identity consistency of the generated transition audio. Then, we adopt Whisper~\cite{radford2023robust} to restrict the word error rate, thereby ensuring the accuracy and clarity of the speech content. In this fashion, the synthesized transition audio realizes controlled duration while the natural smooth tonality is well preserved. The datasets will be released in the future.
More details are provided in the Section \ref{sec:experiment_data}.

\subsection{Problem Formulation}
Given a sequence of audio signal $A = \left \{ a_{1},...,a_{N} \right \}$ as input to model $\mathcal{G}$, our goal is to generate vivid and emotional 3D co-speech gestures $P = \left \{ p_{1},...,p_{N} \right \}$ of the upper human body. $N$ represents the number of synthesized postures corresponding to the audio $A$. Each $p_{i}$ is denoted as $J$ joints with 3D representation. In particular, we define the audio signal as consisting of a head speech, a dynamically inpainted transition speech, and a tail speech. Here, the head and tail speeches are randomly selected from previous co-speech gesture datasets~\cite{liu2022learning, liu2022beat} with a single emotion label, respectively. Following the conventions of previous works~\cite{yoon2020speech,liu2022learning,zhu2023taming,liu2022beat}, we invoke $M$ frame poses as the initial seed to guide generation. The overall objective is expressed as
\begin{equation}
\underset{\mathcal{G}}{\argmin} \left \| P-\mathcal{G} \left ( A, \left \{ p_{1},...,p_{M}\right \}  \right ) \right \|.
\label{Eq1}
\end{equation}
Note that only the generated gestures with head and tail speeches are supervised with ground truth coming from existing datasets~\cite{liu2022learning, liu2022beat}. The transition gestures with $L$ frames, where $L\ll N$, will be weakly supervised through the following processes within a motion transition infusion mechanism and an emotion mixture strategy. The audio signal is fed into an audio encoder for feature extraction. Our overall workflow is shown in Figure~\ref{fig:pipeline}.

\subsection{Weakly-supervised Emotion Transition}
\noindent \textbf{Motion Transition Infusion Mechanism:}
To ensure the temporal coherence of the transition gestures \wrt head/tail ones, we propose a motion transition infusion mechanism to model the temporal association representation between different emotional gestures. As depicted in Figure \ref{fig:pipeline} (left), the total sequential features $\mathbf{f}^{total}$ consist of the features of head $\mathbf{f}^{head}$, transition $\mathbf{f}^{tran}$, and tail $\mathbf{f}^{tail}$. Inspired by \cite{mao2022weakly}, we nominate a head chunk $\mathbf{f}_{chunk}^{head}$ composed of the last $L$ frames of head gesture embeddings and a tail chunk $\mathbf{f}_{chunk}^{tail}$ consisting of the first $L$ ones of the tail gesture embeddings. Here, the dimension of each frame representing gesture embedding is $\mathbb{R}^{1\times D}$. 

In particular, we first calculate the temporal correlation matrix $\mathbf{C}_{trans}^{head}\in \mathbb{R} ^{L \times L}$ between the head chunk embedding and transition embedding.
Here, the temporal correlation matrix represents the temporal variations in the gestures from the head to the transition. Similarly, we obtain the correlation matrix $\mathbf{C}_{trans}^{tail}$ from tail chunk embedding and transition embedding. Then, the global temporal dependency from head to tail is computed via matrix multiplication between $\mathbf{C}_{trans}^{head}$ and $\mathbf{C}_{trans}^{tail}$.
Further, we develop a motion encoder to acquire the sequence-aware style guidance representation based on the global temporal dependency. The style guidance representation is exploited to boost the transition embeddings via an adaptive instance normalization (AdaIN) layer~\cite{huang2017arbitrary}. By doing so, we derive the transition $\mathbf{f}^{tran}$ as
\begin{equation}
\mathbf{f}^{tran}  \!=\! \text{AdaIN}\left \{ \mathbf{f}^{tran} , \text{Enc}(\mathbf{C}_{trans}^{head} \otimes \mathbf{C}_{trans}^{tail})\right \},
\label{Eq2}
\end{equation}
where $\otimes$ indicates matrix multiplication, and \text{Enc} denotes the motion encoder.

\noindent \textbf{Emotion Mixture Strategy:}
Considering obtaining the realistic 3D pose annotations corresponding to dynamically inpainted transition audio is quite difficult, we design an emotion mixture strategy to provide weak supervision of the generated gestures. Our key insight is built upon the fact that different emotions in the head/tail would lead to different gesture motions, thereby the emotion represented by transition gestures would be a mixture of the head and tail ones. As shown in Figure \ref{fig:pipeline} (right), we utilize two learnable parameters as soft emotion labels of the transition gestures. 

Specifically, through computing the correlation matrix between the embeddings of head gestures and tail gestures, we obtain the motion deformation representation $S$ from one emotion to another (\eg, neutral-to-anger). This motion deformation is formulated as:
\begin{equation}
\mathbf{S}_{head}^{tail} \! = \! \text{Softmax}( \mathbf{f}^{head} \otimes {\mathbf{f}^{tail}}^{'}),
\label{Eq3}
\end{equation}
where $'$ indicates the transpose operation. With the help of the motion deformation representation, we obtain the two learnable emotion embeddings $(\mathbf{e}^{head}, \mathbf{e}^{tail})$
from the head and tail, respectively, as
\begin{align}
\mathbf{e}^{head} & = \text{MaxPool} \left \{ \mathbf{W}_{linear} (\mathbf{f}^{head}) \odot \mathbf{S}_{head}^{tail} \right \}, 
\notag
\\
\mathbf{e}^{tail} & = \text{MaxPool} \left \{ \mathbf{W}_{tail} (\mathbf{f}^{head}) \odot \mathbf{S}_{head}^{tail} \right \},
\label{eq4}
\end{align}
where $\mathbf{W}_{linear}$ denotes linear matrix, and $\odot$ means dot product. \text{Maxpool} indicates the AdaptiveMaxPooling operation. $\mathbf{e}^{tail}$ is calculated in a similar way. Then, we obtain the two learnable parameters represented by emotion weights of head and tail gestures, as follows
\begin{equation}
(\sigma, 1-\sigma) = \text{Softmax}(\mathbf{e}^{head}, \mathbf{e}^{tail}).
\label{Eq5}
\end{equation}
Once we acquire the learnable emotion weights, we leverage a pre-trained pose-based emotion classifier to provide weak supervision. In this fashion, the authority of generated transition gestures is well-preserved. The training details of the pre-trained pose-based emotion classifier are provided in the supplementary material.

\vspace{0.5em}
\noindent \textbf{Keyframe Posture Sampling:}
Conventionally, researchers~\cite{liu2022learning, zhu2023taming, yoon2020speech} directly leverage the padded initial poses as conditional seeds to guide the co-speech gesture generation. In long-sequence modeling, an intuitive manner is to extend the initial pose length $M$ for adapting the overall synthesized gestures' length. However, this would lead to the poor generalizability of the network (\ie, the performance gains degradation from the reduction of initial poses). Therefore, we propose a simple yet effective VAE-based~\cite{sohn2015learning} keyframe sampler to provide high-fidelity posture prior conditions while enabling diverse results. Evenly, we split the annotated sequence into several chunks keeping the same length with the transition of length $L$. The sampler is trained with the keyframe randomly selected to reconstruct the corresponding chunk. In the inference phase, the keyframe sampler samples diverse chunks, blending as the initial postures for gesture generation. Since the transition sequence lacks the pose annotation, we randomly select the frame from head or tail sequences for posture sampling.

To further enhance the sequence-aware correspondence of the generated co-speech gestures, we leverage the diverse initial postures as the query $Q$ to match the key features $K$ and value features $V$ in the transformer-based backbone~\cite{vaswani2017attention}. Similar to \cite{qi2023diverse}, we adopt a motion discriminator to ensure the temporal smoothness of the generated results. For more details about network architecture please refer to supplementary material.

\vspace{0.5em}
\subsection{Objective Functions}
\vspace{0.5em}
\noindent{\textbf{Reconstruction Loss:}}
We leverage the ground truth 3D pose annotation of the head and tail to constrain the generated co-speech gestures as:
\begin{align}
\mathcal{L}_{rec} \! & = \! \left \| P_{\left \{ head, tail \right \} }  \!-\! \hat{P}_{\left \{ head, tail \right \}}\right \| _{1},
\label{Eq6}
\end{align}
where $\hat{P}_{\left \{ head, tail \right \}}$ denotes generated gestures of head and tail speeches.

\vspace{0.5em}
\noindent{\textbf{Adversarial Learning Loss:}}
To ensure the realism of the generated gestures, we further exploit the adversarial training loss, expressed as:
\begin{align}
\mathcal{L}_{adv} & = \mathbb{E} _{P} \left [ \log \mathcal{D}(P) \right ] 
\notag 
\\ &+ \mathbb{E} _{A} \left [ \log(1- (\mathcal{G}\left ( A , \left \{ p_{1},...,p_{M}\right \}  \right )) \right ],
\label{eq7}
\end{align}
where $\mathcal{D}$ denotes the motion discriminator and $\mathcal{G}$ means gesture generator. 

\vspace{0.5em}
\noindent{\textbf{Weakly Supervision Loss:}}
We leverage the pre-trained pose-based emotion classifier to provide weak supervision of the transition gestures upon the learnable emotion weights:
\begin{align}
\mathcal{L}_{emotion}  = -y \log \mathcal{F}(\hat{P}_{trans}),
\label{Eq8}
\end{align}
where $y$ is the learnable emotion label, $\mathcal{F}$ is the emotion classifier, $\hat{P}_{trans}$ is the generated transition gestures. 

\noindent Finally, the overall objective is:
\begin{align}
\!\!\min_{G}\max_{D}\mathcal{L}_{total} \!=\!  \lambda_{r} \mathcal{L} _{rec} + \lambda_{adv} \mathcal{L} _{adv} + \mathcal{L} _{emotion}.
\label{eq9}
\end{align}
The $\lambda_{r}$, and $\lambda_{adv}$ are weight coefficients.

\section{Experiments}
\label{sec:experiments}

\begin{table*}[t]
\centering
\caption{Statistics comparison of existing 3D co-speech gesture datasets with ours. Our \textbf{BEAT-ETrans} and \textbf{TED-ETrans} are built upon the existing BEAT~\cite{liu2022beat} and TED-Expressive~\cite{liu2022learning}, respectively. To the best of our knowledge, we are the first to present two large datasets with emotion transition human speech. }
\label{tab:dataset}
\footnotesize
\resizebox{\textwidth}{!}{%
\begin{tabular}{llcccccccc}
\toprule
                            &                                                                              & \multicolumn{7}{c}{Modality}                                                                                                                                                              &                                    \\ \cmidrule(r){3-9}
\multirow{-2}{*}{Dataset}   & \multirow{-2}{*}{\begin{tabular}[c]{@{}c@{}}Joint\\ Annotation\end{tabular}} & Body        & Hand        & Audio                               & Text                                & Speakers       & Single Emotion             & \textbf{Emotion Transition}                  & \multirow{-2}{*}{Duration (hours)} \\ \midrule \midrule
TED~\cite{yoon2020speech}                         & pseudo label                                                                 & 9           & \ding{55}           & \ding{51}          & \ding{51}          & 1,766          & \ding{55} & \ding{55}          & 106.1                              \\
SCG~\cite{habibie2021learning}                         & pseudo label                                                                 & 14          & 24          & \ding{51}          & \ding{55}          & 6              & \ding{55} & \ding{55}          & 33                                 \\
Trinity~\cite{ferstl2018investigating}                     & mo-cap                                                                       & 24          & 38          & \ding{51}          & \ding{51}          & 1              & \ding{55} & \ding{55}          & 4                                 \\
ZeroEGGS~\cite{ghorbani2023zeroeggs}                    & mo-cap                                                                       & 27          & 48          & \ding{51}          & \ding{51}          & 1              & \ding{55} & \ding{55}          & 2                                 \\
BEAT~\cite{liu2022beat}                        & mo-cap                                                                       & 27          & 48          & \ding{51}          & \ding{51}          & 30             & \ding{51} & \ding{55}          & 35                                 \\
TED-Expressive~\cite{liu2022learning}              & pseudo label                                                                 & 13          & 30          & \ding{51}          & \ding{51}          & 1,764          & \ding{55} & \ding{55}          & 100.8                              \\ \midrule
\rowcolor[HTML]{ECF4FF} 
\textbf{BEAT-ETrans (ours)} & \textbf{mo-cap}                                                              & \textbf{27} & \textbf{48} & \textbf{\ding{51}} & \textbf{\ding{51}} & \textbf{30}    & \textbf{8}                 & \textbf{\ding{51}} & \textbf{161.3}                     \\
\rowcolor[HTML]{ECF4FF} 
\textbf{TED-ETrans (ours)}  & \textbf{pseudo label}                                                        & \textbf{13} & \textbf{30} & \textbf{\ding{51}} & \textbf{\ding{51}} & \textbf{1,764} & \textbf{6}                 & \textbf{\ding{51}} & \textbf{59.8}                      \\ \bottomrule
\end{tabular}
}
\end{table*}

\begin{table*}[t]
\centering
\caption{Comparison with the start-of-the-art methods on our newly collected BEAT-ETrans and TED-ETrans datasets.
$\uparrow$ denotes the higher the better, and $\downarrow$ indicates the lower the better. $\pm$ means 95\% confidence interval. 
}
\label{tab:table1}
\footnotesize
\setlength{\tabcolsep}{2.5 mm}
\begin{tabular}{lcccccccc}
\toprule
\multirow{2}{*}{Models} & \multicolumn{4}{c}{BEAT-ETrans}               & \multicolumn{4}{c}{TED-ETrans}               \\ \cmidrule(r){2-5} \cmidrule(r){6-9}
                        & FGD$_{h+t}$  $\downarrow$ & FGD$_{trans}$ $\downarrow$ & BC $\uparrow$   & Diversity $\uparrow$    & FGD$_{h+t}$ $\downarrow$  & FGD$_{trans}$ $\downarrow$ & BC $\uparrow$   & Diversity $\uparrow$  \\ \midrule \midrule
Seq2Seq~\cite{yoon2019robots}\textcolor[HTML]{C0C0C0}{$_{ICRA'19}$}                 & 40.95 & 47.93          & 0.141 & 96.66$^{\pm 2.16}$  & 29.60 & 49.47          & 0.265 & 72.81$^{\pm1.99}$ \\
S2G~\cite{ginosar2019learning}\textcolor[HTML]{C0C0C0}{$_{CVPR'19}$}                     & 25.56 & 37.04          & 0.671 & 98.26$^{\pm2.04}$  & 18.16 & 41.63          & 0.824 & 76.82$^{\pm2.32}$ \\
Trimodal~\cite{yoon2020speech}\textcolor[HTML]{C0C0C0}{$_{TOG'20}$}                & 14.09 & 42.50          & 0.764 & 100.87$^{\pm2.12}$ & 21.06 & 33.20          & 0.758 & 82.87$^{\pm1.86}$ \\
CAMN~\cite{liu2022beat}\textcolor[HTML]{C0C0C0}{$_{ECCV'22}$}                   & 9.03  & 27.53          & 0.794 & 118.46$^{\pm2.33}$ & 19.28 & 41.04          & 0.785 & 79.03$^{\pm1.49}$ \\
HA2G~\cite{liu2022learning}\textcolor[HTML]{C0C0C0}{$_{CVPR'22}$}                   & 7.28  & 25.79          & 0.779 & 121.77$^{\pm2.31}$ & 16.72 & 40.38          & 0.787 & 80.14$^{\pm1.65}$ \\
DiffGesture~\cite{zhu2023taming}\textcolor[HTML]{C0C0C0}{$_{CVPR'23}$}             & 6.68  & 25.03          & 0.788 & 122.29$^{\pm2.01}$ & 18.69 & 25.13          & 0.818 & 92.01$^{\pm2.07}$ \\ \midrule
\rowcolor[HTML]{ECF4FF} 
\textbf{Ours} & \textbf{4.42} & \textbf{18.84} & \textbf{0.881} & \textbf{124.93$^{\pm 2.10}$} & \textbf{12.19} & \textbf{23.54} & \textbf{0.906} & \textbf{93.79$^{\pm 2.53}$} \\ 
\bottomrule
\end{tabular}
\end{table*}

\subsection{Datasets and Experimental Setting}
\label{sec:experiment_data}

\noindent{\textbf{BEAT Emotion Transition Dataset (BEAT-ETrans):}}
Since there are only \textbf{\emph{single emotion}} labels of aligned speech-gesture corpus in the original BEAT dataset~\cite{liu2022beat}, to satisfy our insight on emotion transition co-speech gesture generation modeling, we recollect a BEAT Emotion Transition Dataset (dubbed BEAT-ETrans). In particular, we resample the motion FPS as $15$ and intercept the continuous $60$ frames with stride $30$ as the head/tail clips. Here, the head clips are all annotated as \emph{neutral}, and tail clips are denoted with the other seven emotions: \emph{anger, happiness, fear, disgust, sadness, contempt, and surprise}. As for one head speech, we randomly select two or three tails with different emotions to construct the head-tail pairs.
By leveraging the two-second (\ie, corresponding $30$-frame postures) transition to blend the heads and tails, we obtain the $10$-second human speech clips in our BEAT-ETrans.  
We obtain $58,077$ clips, including a total of $161.3$ hours reported in Table \ref{tab:dataset}. Then the clip numbers of training/validation/testing sets are randomly split as $41,908$/$4,077$/$12,092$. In all of our experiments, we utilize the upper body with $71$ joints.

\vspace{0.5em}
\noindent{\textbf{TED Emotion Transition Dataset (TED-ETrans):}} Inspired by \cite{liu2022learning, qi2023emotiongesture}, we further newly collect a TED Emotion Transition Dataset (dubbed TED-ETrans) based on more than 1.7K speakers from in-the-wild TED talk show videos, demonstrated in Table \ref{tab:dataset}. Due to the lack of emotional labels, we first leverage the annotated BEAT dataset to pre-train an audio-based emotion classifier for labeling TED audios. To ensure the authority of emotion labels, we set the classification threshold as $\ge 0.95$, and the two uncommon emotions (\ie \emph{fear, disgust}) are dropped. Then we maintain the same data pre-processing strategy with our BEAT-ETrans to obtain a total of $21,515$ clips with $59.8$ hours. The final clip division criteria of the TED-ETrans dataset are training/validation/testing with 15,061/2,152/4,302 respectively. In practice, the $43$ upper body joints are leveraged in the experiments. 

\vspace{0.25em}
\noindent{\textbf{Implementation Details:}} 
We set the total generated co-speech gesture length as $N=150$, and the transition and chunk lengths are $L=30$. Conventionally, we leverage $M=4$ frames as the reference initial poses. The feature dimension $D=512$ in practice. The raw audio of human speech is converted to mel-spectrograms with FFT window size $1024$, and hop length $512$. The audio encoder takes the ResNetSE34~\cite{chung2020defence} as the backbone. Empirically, we set $\lambda_{r}=20$, and $\lambda_{adv} = 2$. Our models are implemented on the Pytorch platform with a single NVIDIA Tesla V100 GPU. The initial learning rate is set to $0.0003$ by utilizing Adam Optimizer. The whole training takes $100$ epochs with a batch size of $96$.

\begin{table*}[t]
\centering
\footnotesize
\caption{Ablation study on different components of the proposed method. \ding{51} indicates the employment of a certain module.
$\uparrow$ denotes the higher the better, and $\downarrow$ indicates the lower the better. $\pm$ means 95\% confidence interval. MTIM: Motion Transition Infusion Mechanism; EMS: Emotion Mixture Strategy; FKS: Keyframe Sampler.}
\label{tab:table2}
\setlength{\tabcolsep}{2 mm}
\begin{tabular}{cccc|cccccccc}
\toprule
\multicolumn{4}{c|}{Model Variations}                                                                                       & \multicolumn{4}{c}{BEAT-ETrans}                                                                                                     & \multicolumn{4}{c}{TED-ETrans}                                                                                                      \\ \cmidrule(r){1-4} \cmidrule(r){5-8} \cmidrule(r){9-12}
Baseline                   & MTIM                       & EMS                        & KFS                        & FGD$_{h+t}$  $\downarrow$                          & FGD$_{trans}$ $\downarrow$                & BC $\uparrow$                           & Diversity $\uparrow$                           & FGD$_{h+t}$  $\downarrow$                           & FGD$_{trans}$ $\downarrow$               & BC $\uparrow$                           & Diversity $\uparrow$                          \\ \midrule \midrule
\ding{51} &                            &                            &                            & 21.21                        & 56.33                         & 0.701                         & 88.89$^{\pm1.58}$                          & 26.44                         & 47.87                         & 0.712                         & 74.19$^{\pm2.75}$                         \\
\ding{51} & \ding{51} &                            &                            & 10.75                        & 27.94                         & 0.866                         & 108.69$^{\pm2.63}$                         & 20.95                         & 36.38                         & 0.827                         & 82.42$^{\pm2.24}$                         \\
\ding{51} & \ding{51} & \ding{51} &  & 5.69                         & 21.70                         & 0.878                         & 112.80$^{\pm1.95}$                         & 14.02                         & 26.21                         & 0.900                         & 87.21$^{\pm2.12}$                         \\\rowcolor[HTML]{ECF4FF} 
\ding{51} & \ding{51} & \ding{51} & \ding{51} & \textbf{4.42} & \textbf{18.84} & \textbf{0.881} & \textbf{124.93$^{\pm2.10}$} & \textbf{12.19} & \textbf{23.54} & \textbf{0.906} & \textbf{93.79$^{\pm2.53}$} \\
\bottomrule
\end{tabular}
\end{table*}

\vspace{0.25em}
\noindent{\textbf{Evaluation Metrics:}} 
To fully evaluate the realism and diversity of the generated co-speech gestures, we introduce various metrics:
\vspace{-0.2em}
\begin{itemize}[leftmargin=*]
    \item \textbf{FGD}: Fréchet Gesture Distance (FGD) \cite{yoon2020speech} is utilized to measure the distribution distance between the realistic sequential gestures and generated ones. Since we only have the 3D joint annotation of the head/tail, we take the network architecture provided in \cite{yoon2020speech,liu2022learning} to train the auto-encoder for distance computing on the two datasets, respectively. The FGD of transition gestures is calculated as the average value between the distribution distance of transition and head/tail, indicated as FGD$_{trans}$. Similarly, FGD$_{h+t}$ means the distance between the generated head/tail gestures and ground truth.
    \item \textbf{BC}: Beat Consistency Score (BC) \cite{liu2022beat, liu2022learning} measures the speech audio alignment degree with the generated co-speech gestures.
    \item \textbf{Diversity}: Similar to \cite{liu2022learning, zhu2023taming}, we exploit the same feature extractor in FGD to obtain the feature embeddings of the generated gestures. The diversity reflects the average distance between $500$ random combination pairs in the testing set of $60$ speech audios.
\end{itemize}

\subsection{Quantitative Evaluation}
\vspace{-0.2em}
\noindent{\textbf{Comparisons with SOTA Methods:}} To the best of our knowledge, we are the first to explore the co-speech gesture generation with emotion transition human audios. To fully verify the superiority of our method, we implement various state-of-the-art single-emotion-based counterparts: Seq2Seq~\cite{yoon2019robots}, S2G~\cite{ginosar2019learning}, Trimodal~\cite{yoon2020speech}, CAMN~\cite{liu2022beat}, HA2G~\cite{liu2022learning}, and DiffGesture~\cite{zhu2023taming}. For a fair comparison, all the models are implemented by the source code released by the authors. Note that since we only have the ground truth of head and tail gestures, several recent VAE-based works~\cite{ao2023gesturediffuclip, zhi2023livelyspeaker, yi2023generating, ao2022rhythmic} cannot be directly applied in the experiments (or they have not released codes so far).

As reported in Table \ref{tab:table1}, we adopt the FGD$_{h+t}$,  FGD$_{trans}$, BC, and Diversity for a well-rounded view of comparisons. Our method outperforms all the competitors by a large margin on both two datasets. Remarkably, on the TED-ETrans dataset, our method even achieves 34.8\% (\ie, $(18.69-12.19)/18.69 \approx 34.8\%$) improvement over the sub-optimal counterparts in FGD$_{h+t}$. We observe both the DiffGesture~\cite{zhu2023taming} and ours synthesize the high-fidelity gestures of head/tail speech with much lower FGD$_{h+t}$ than others. However, the DiffGesture shows worse performance on FGD$_{trans}$ due to lack of supervision. In terms of diversity, our simple yet effective keyframe sampler provides authority and diverse initial postures as the reference, thus enabling us to demonstrate diverse gesture styles compared to other counterparts. Moreover, we find that the diversity scores on BEAT-ETrans are much higher than those on TED-ETrans dataset. This can be attributed to the more complex human joints in the BEAT-ETrans dataset.

\vspace{0.5em}
\noindent{\textbf{Ablation Study:}}
To further verify the effectiveness of our proposed methods, we conduct the ablation study of different components as variations, reported in Table ~\ref{tab:table2}. The baseline model is implemented by a simple transformer-based pipeline with stacking three times blocks in the encoder-decoder. Obviously, all the combinations of our proposed components have positive impacts on the generated results. Specifically, by adding the motion transition infusion mechanism to the baseline, the indicator BC has achieved significant improvement (\eg, $0.701$ $\to$ $0.866$ in the BEAT-Etrans). This result verifies that our motion transition infusion mechanism effectively models the temporal correlation between the transitions \wrt head/tail gestures, thus leading to the generated results preserving rhythm coherency with given speech, globally.

Moreover, adopting the emotion mixture strategy ideally improves the performance of FGD$_{trans}$ on both two datasets. This indicates that the learnable emotion mixture wights can provide effective weak supervision by leveraging the pre-trained pose-based emotion classifier. Besides, we have observed significant improvement in the performance of FGD$_{h+t}$ during this phase compared to the previous version. This aligns with our transformer backbone's emphasis on modeling the sequential temporal correlations as a whole. The better transition gestures encourage our model to maintain better temporal consistency, thereby the head/tail gestures achieve better results.

Finally, after additionally employing the keyframe sampler to produce authority initial postures as the reference, our method obtains the best performance. Although the properties of BC and FGD on both datasets just have slightly improvement, the diversity realizes a noticeably better achievement (\eg, 112.80 $\to$ 124.93 in the BEAT-ETrans dataset). This highly supports our insight into keyframe-driven diversification strategy.

\begin{figure*}[t]
\begin{center}
\includegraphics[width=0.9\linewidth]{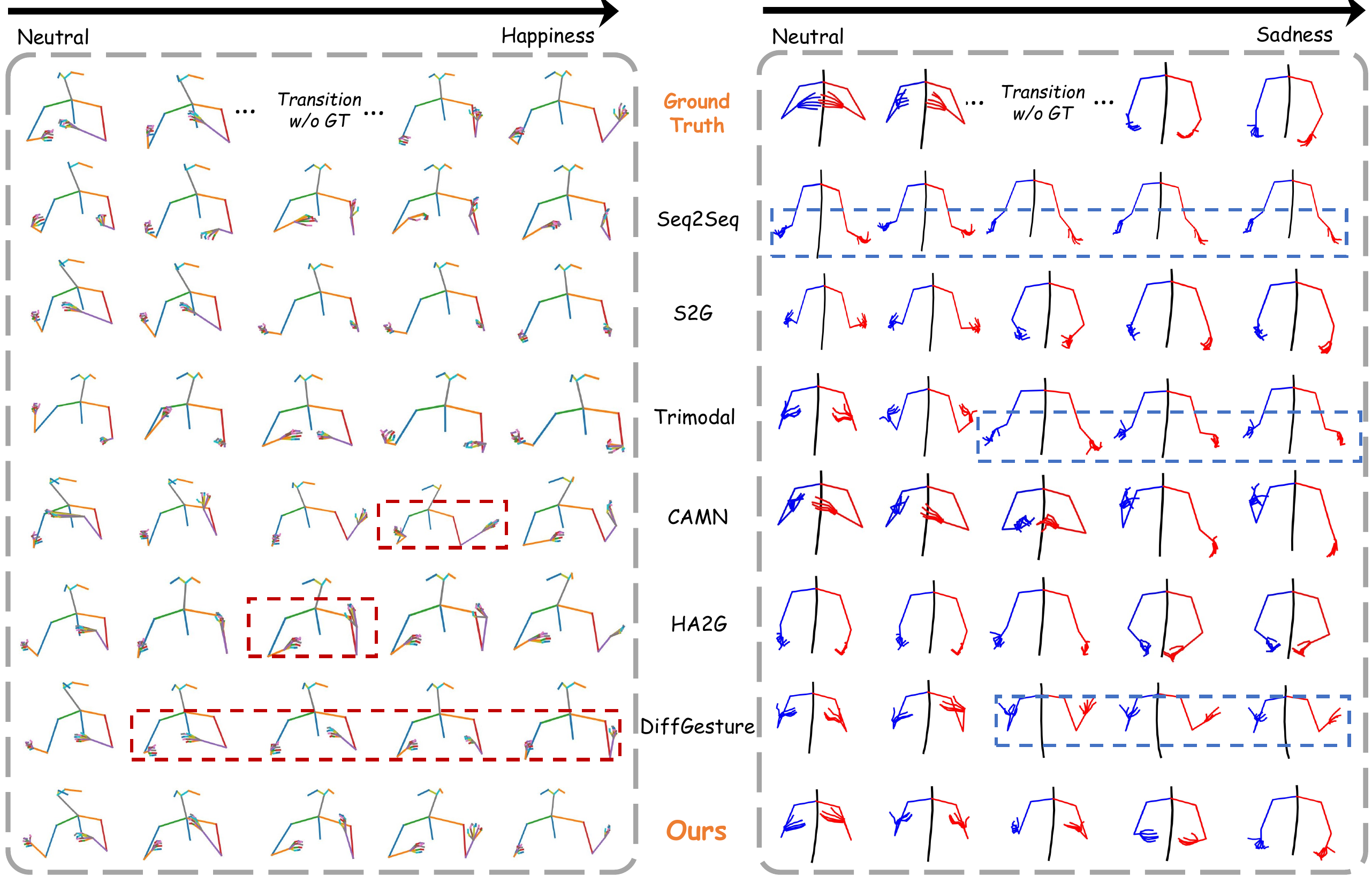}
\end{center}
\vspace{-1em}
\caption{Visualization of our generated 3D co-speech gestures against various state-of-the-art methods. The samples of the left part are from our newly collected TED-ETrans dataset, and the samples of the right part are from our BEAT-ETrans dataset. Best view on screen.
}
\label{fig:comparison}
\vspace{-0.5em}
\end{figure*}

\subsection{Qualitative Evaluation}

\noindent{\textbf{Visualization:}}
To fully demonstrate the performance of our method, we show the visualized keyframes generated from ours compared with counterparts on our newly collected TED-ETrans and BEAT-ETrans datasets, respectively. As depicted in Figure \ref{fig:comparison}, our method displays vivid and diverse results against others. In particular, we observe that Seq2Seq and Trimodal tend to synthesize unreasonable and stiff results (\eg, the blue rectangle in the right BEAT-ETrans dataset). Although CAMN and HA2G can generate natural upper-body postures, we find that they sometimes produce unreliable subtle fingers (\eg, the red rectangle in the left TED-ETrans dataset). Both our method and DiffGesture create reasonable gestures. However, the results synthesized by DiffuGesture are mismatched with the emotion transitions. In contrast, our method can synthesize the synchronous motions (\eg, in the BEAT-ETrans, the arms become droopy as emotion turns to sadness). Meanwhile, we further verify the diversification results as shown in Figure \ref{fig:teaser}. Given the same input audio, our method generates diverse and vivid co-speech gestures.
Please refer to the supplementary material for more visualization results.

\begin{figure}[t]
\begin{center}
\includegraphics[width=0.92\linewidth]{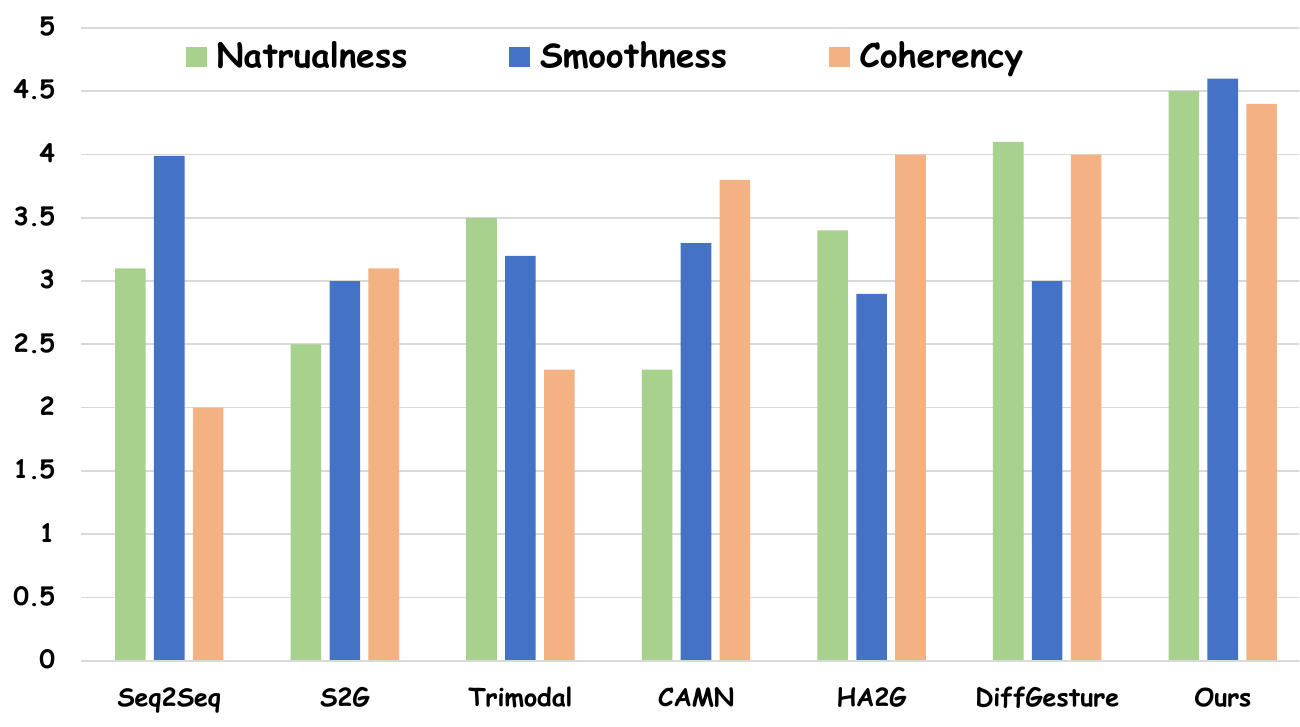}
\end{center}
\vspace{-1em}
\caption{User study on gesture naturalness, motion smoothness, and speech-gesture coherency. 
}
\label{fig:user_study}
\vspace{-1.5em}
\end{figure}

\vspace{0.5em}
\noindent{\textbf{User Study:}}
To further analyze the quality of results by various counterparts and ours, we conduct a user study by visualizing the results and inviting $15$ subjects. For each emotion in both datasets, we randomly select 12 samples for each participant (7 emotions in BEAT-ETrans, 5 emotions in TED-ETrans). All the participants are recruited anonymously from schools including various majors. In particular, the subjects are required to rate the generated co-speech gestures from $0$ to $5$ (the higher, the better) in terms of naturalness, motion smoothness, and speech-gesture coherency. The results are demonstrated in Figure \ref{fig:user_study}.
Our method showcases the best performance compared with all the competitors. Especially in terms of motion smoothness, our method achieves noticeable advantages, indicating the effectiveness of our proposed motion transition infusion mechanism and the emotion mixture strategy.

\section{Conclusion}
In this paper, we introduce a new task of 3D co-speech gesture generation given emotion transition human speech. We therefore newly collected two datasets named the BEAT-ETrans and the TED-ETrans to fulfill this goal while significantly facilitating the research on 3D human motion modeling. 
Then, we fully take advantage of the sequential temporal correlation via a motion transition infusion mechanism to ensure the generated gestures preserve temporal coherence. Furthermore, we design an emotion mixture strategy to supply emotional weak supervision of the synthesized transitions. Extensive experiments conducted on our two newly collected datasets show the superiority of the method. As our method intends to generate diverse and vivid emotion transition gestures, we will investigate diversifying the 3D gesture with temporal smooth sampling, instead of the keyframe-wise manner.

\vspace{0.5em}
\noindent{\textbf{Acknowledgements.}} This work is funded in part by the Theme-based Research Scheme grant (No.T45-205/21-N) and the InnoHK funding, Hong Kong SAR.

\clearpage
\setcounter{page}{1}

\twocolumn[{%
\renewcommand\twocolumn[1][]{#1}%
\maketitlesupplementary
\begin{center}
    \centering
    \includegraphics[width=0.98\linewidth]{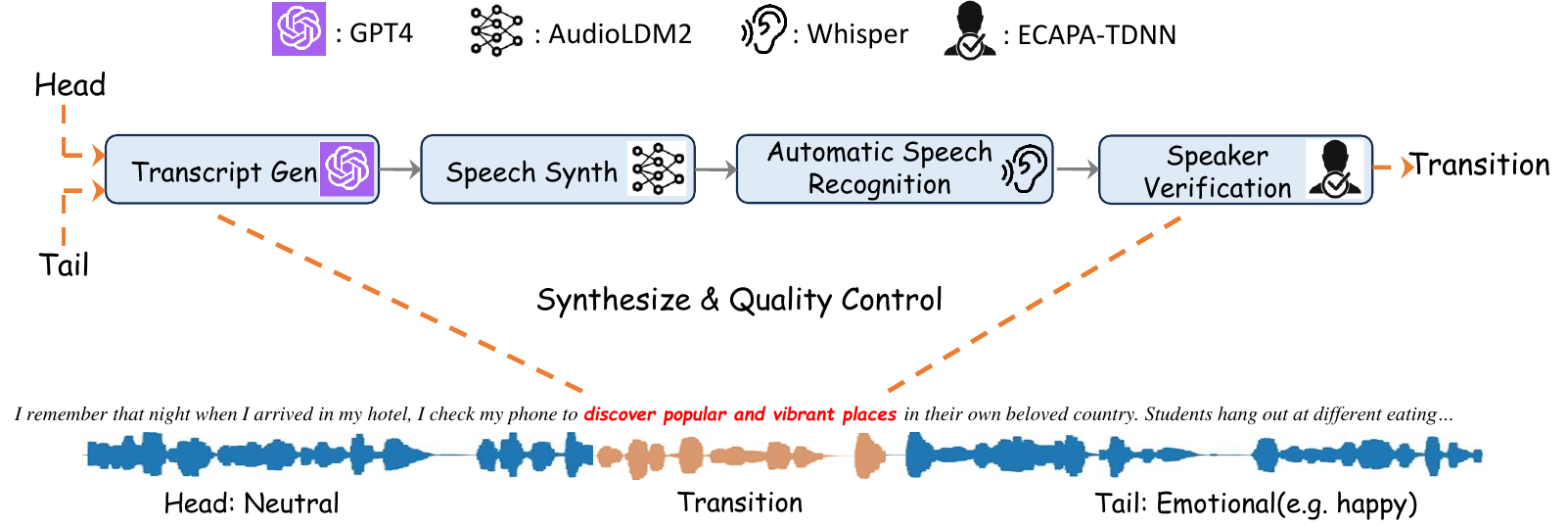}
    \captionof{figure}{The pipeline of dataset construction. Head and tail audios as well as the corresponding transcripts are fed into the pipeline to generate a smooth and high quality transition. }
    \label{fig:dataset}
\end{center}
}]

\section{Overview}
To demonstrate the effectiveness of our data construction techniques and the proposed method of emotion transition co-speech gesture generation, we further elaborate on the detailed data synthesis and vision perception in the supplementary material. The additional content is illustrated in the following folds:
\begin{itemize}[leftmargin=*]
    \item Dataset Construction
    \item Architecture Details
    \item Additional Experiments
\end{itemize}

\section{Dataset Construction}
\label{sec:dataset}
\textbf{\emph{We will release our newly collected the TED-ETrans and BEAT-ETrans datasets in the future.}}
The overall pipeline of our approach to constructing the dataset is displayed in Figure \ref{fig:dataset}. The details involve the following steps:

\begin{figure}
\begin{center}
\includegraphics[width=0.97\linewidth]{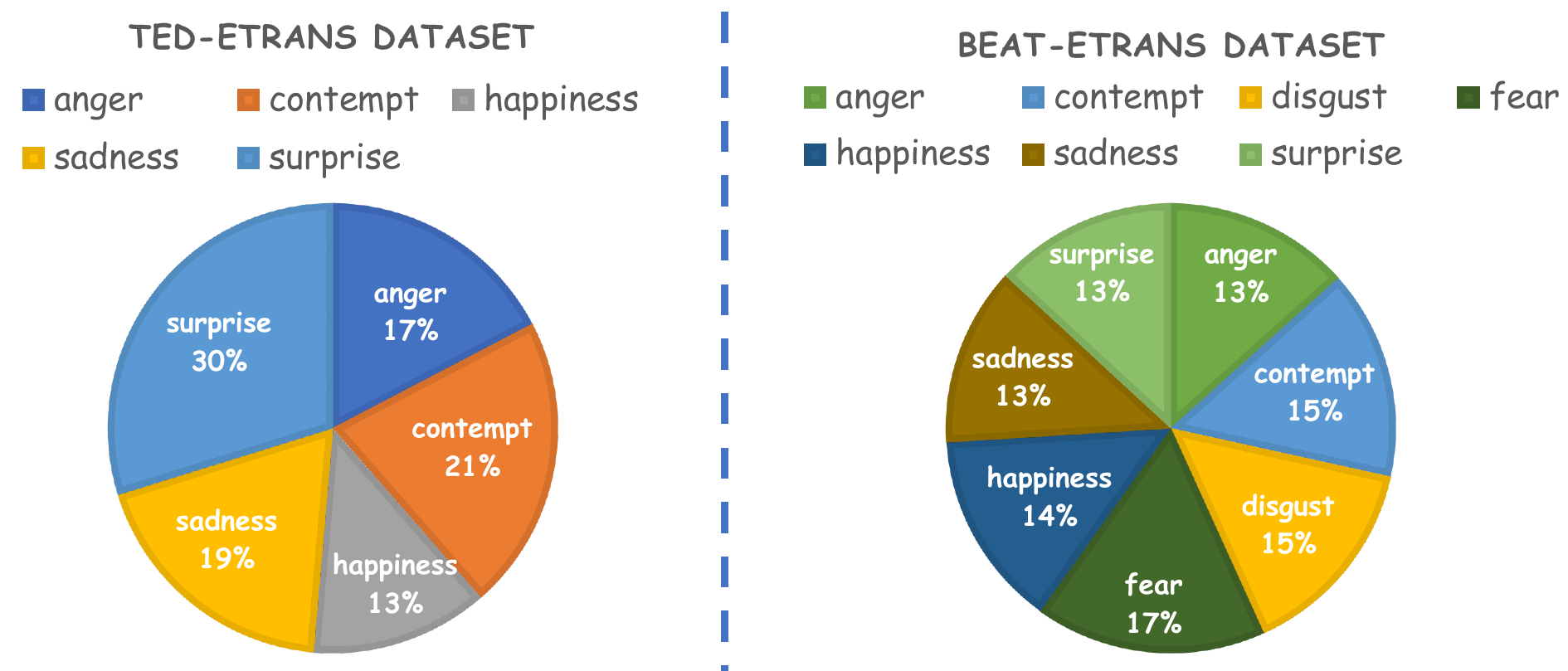}
\end{center}
\caption{Details of emotion transition distribution of our newly collected TED-ETrans and BEAT-ETrans datasets. All the transitions start from the neutral emotional speeches.}
\label{fig:emotion_dis}
\end{figure}

\paragraph{Segmentation and Emotion Labeling}: We first divide the previously aligned single emotion co-speech gesture datasets~\cite{liu2022beat,liu2022learning} into head and tail segments by splitting the original audio into 4-second clips. Heads are identified as clips with neutral emotions, while tails contain various emotions. This segmentation was achieved using either the pre-annotated dataset's emotion labels or an emotion classifier. Both head and tail segments originated from the same speaker, ensuring vocal tone consistency. 
    
\paragraph{Emotion Transition}: The head segments consistently exhibit neutral emotions, while the tails display a variety of emotional states. In our approach, we intentionally avoided pairing segments with extreme emotional shifts (\eg, happiness-to-anger, happiness-to-sadness). Such drastic transitions are infrequent in natural speech and not only result in less smooth transitions in both speech and textual contexts but also risk introducing a long-tail phenomenon in the dataset. By avoiding these extremes, we aimed to maintain a more balanced and realistic dataset distribution as shown in Figure \ref{fig:emotion_dis}.
    
\paragraph{Transcript Generation with GPT-4}: We engage GPT-4 to generate transitional text between the head and tail clips. The GPT-4 is instructed to create a smooth transition in both content and emotion, producing about 5-10 words. For each data sample, GPT-4 generated three candidate transitions, each accompanied by a confidence score, returned in JSON format. We finally discard samples with low confidence or excessive length. 

\paragraph{Synthesis of Transition Speech}: We employ the AudioLDM2~\cite{liu2023audioldm, liu2023audioldm2} model for audio inpainting, ensuring natural and time-controlled speech synthesis. Speaker embeddings are extracted using SpeechBrain's ECAPA-TDNN to measure the consistency of the transition speech with the head and tail segments. Samples with significant speaker embedding discrepancies are excluded. We ensure the head, tail, and synthesized parts share the same speaker's tone, maintaining consistency.
    
\paragraph{Quality Control through ASR}: We utilize Whisper~\cite{radford2023robust} for automatic speech recognition (ASR) on transition speech. ASR transcripts are compared to ground truth, and samples with the word error rate of over 0.125 are re-synthesized for better accuracy and clarity.

\paragraph{Final Note}: We observe that GPT-3.5 often produces similar candidates, lacking diversity, thus our preference for GPT-4. Our final prompt structure, designed to guide the model in generating contextually and emotionally coherent transitions, is presented below:

\begin{minipage}[t]{0.45\textwidth}
\phantomsection
\vspace{2mm}
    \begin{tcolorbox}[colback=white!95!gray,colframe=gray!50!black,rounded corners,label={template-fsd}, title={Prompt}]
    \begin{lstlisting}[breaklines=true, xleftmargin=0pt, breakindent=0pt, columns=fullflexible, mathescape]
As a skilled playwright, you've been assigned a task to fill in the blanks. You will be given two sentences (in a talk) with distinct emotions, and your job is to provide a transition of **10 words** to ensure a natural emotional and semantic flow between them. For each blank, you should return three potential options along with your confidence level in your responses in JSON format. **DO NOT** return anything else.
JSON template:
{
    "opt1": option 1,
    "opt2": option 2,
    "opt3": option 3,
    "confi": confidence score scale from 1 to 5,
}
	
Input:\n
    \end{lstlisting}
    \end{tcolorbox}
    \label{prompt}
\end{minipage}

We first define GPT-4 prompt and evaluate sentence completeness \textbf{confidence scores} three times to select the best fit for semantic clarity. 
Moreover, we add a manual review on each transcript to drop the unnatural sentences, including 30 English native speakers' evaluation of grammatical correctness/ logical coherence/ clarity of expression. 
We are unable to extract the transitional segments directly from lengthy videos due to the absence of speech recognizers for identifying multiple emotions within a single audio.
Thus, we leverage the advanced LLM GPT-4 combined with the manual effort to construct the natural and smooth emotion transition datasets.

\section{Architecture Details}
\label{sec:architecture}

\begin{figure}[t]
\begin{center}
\includegraphics[width=0.97\linewidth]{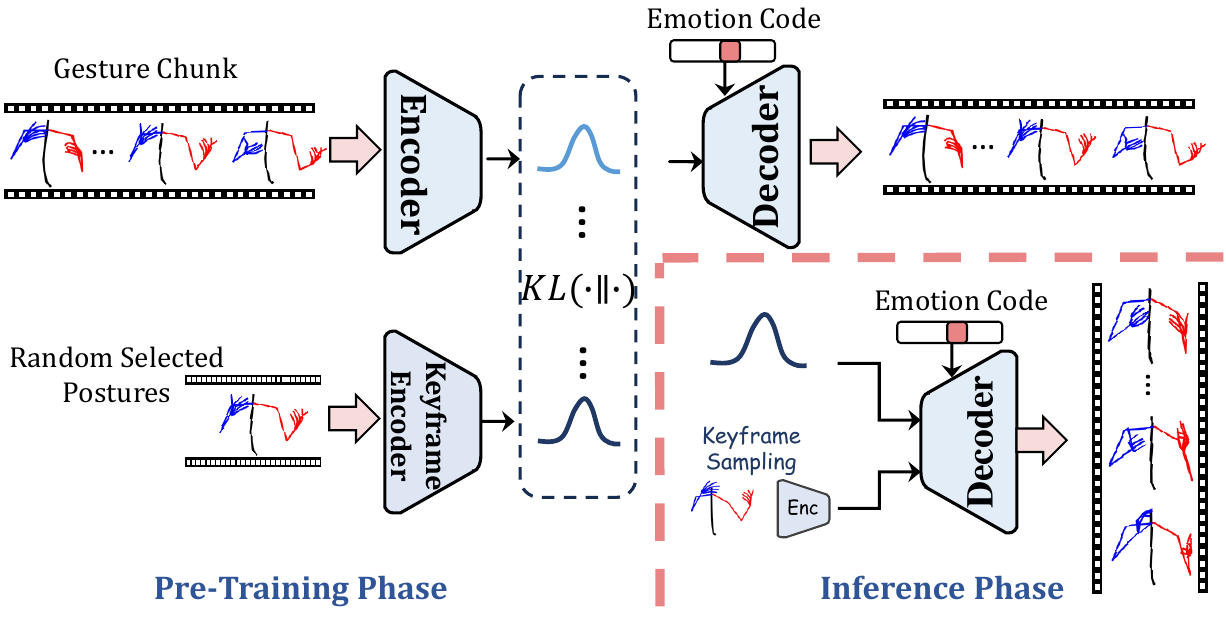}
\end{center}
\caption{Details of our proposed keyframe sampling strategy. Once we obtain the pre-trained keyframe encoder, we leverage it to model the conditional distribution, producing diverse initial postures as the reference.}
\label{fig:keyframe}
\end{figure}

\paragraph{Audio Encoder.} Inspired by \cite{chung2020defence, liu2022learning, zhu2023taming}, the backbone of our audio encoder $E_{a}$ is constructed as ResNetSE34. We adopt three stacking blocks and leverage the 2D-convolution-based header to map the dimension of audio features to be $N\times 512$, where $N$ is the temporal dimension. 

\paragraph{Transformer-based backbone.}
We leverage the diversified authority initial postures to interact with the extracted audio features. In particular, we leverage the pose reference $Q$ to match the key features $K$ and value features $V$ in the transformer-based encoder via three times Multi-Head Attention (MHA) \cite{vaswani2017attention}, expressed as:
\begin{align}
MultiHead(Q, K, V) = softmax(\frac{QK }{\sqrt{d} } )V,
\end{align}
where $d$ is a normalization constant. 

\paragraph{Pose-based Emotion Classifier.} In our emotion mixture strategy, we pre-train a pose-based emotion classifier for providing emotional weak supervision on the generated transition gestures. Specifically, the emotion classifier directly leverages the transformer backbone, the same as the pipeline encoder, to extract the sequential pose features. Then, we utilize an MLP-based classifier header on the pose gestures to produce the final emotion categories. In the BEAT-ETrans dataset, our pre-trained emotion classifier achieves 99.92\% accuracy. In the TED-ETrans dataset, the accuracy is 99.26\%.

\paragraph{Keyframe Sampler.} We design a simple but effective VAE-based keyframe sampler to produce authority initial postures as motion cues, thereby facilitating the diversification of the generated 3D co-speech gestures. As shown in Figure \ref{fig:keyframe}, the keyframe sampler aims to model the conditional distribution upon the given randomly selected postures. In the pre-training phase, the posterior distribution is denoted as the latent variable from the encoded chunk-wise gestures. The prior distribution of this latent variable is modeled by the keyframe encoder. The training goal in this phase is to minimize the distance between the posterior distribution and the prior one via KL divergence represented as $KL(\cdot \parallel \cdot )$ in Figure \ref{fig:keyframe}. Meanwhile, we exploit the $L_{1}$ loss to constrain the reconstructed chunk-wise gestures.

\section{Additional Experiments}
\label{sec:architecture}
\subsection{Metric Calculation Details}
Inspired by \cite{liu2022learning,yoon2020speech}, we take FGD to evaluate whether the generated gestures maintain realism with the ground truth ones in the perceptive of distribution. Conventionally, the feature extractor of FGD is calculated to embed overall sequential gestures into latent space and then utilize a decoder for reconstruction. However, since we do not have the ground truth of the transition gestures, we newly pre-train the feature extractor with the transition length $L$. In the inference stage, FGD$_{h+t}$ is calculated by averaging the distances between five randomly selected chunks of length $L$ from the head/tail and GT, respectively. Similarly, FGD$_{trans}$ is computed as the average value between the distance of transition and five randomly selected chunks of head/tail. \textbf{\emph{We will release the code of our pipeline and evaluation metrics in the future.}}

\subsection{Additional Ablation Study Experiments}

As reported in Table~\ref{tab:ablation_adversarial}, after adding the adversarial loss, $FGD_{trans}$ and BC achieve better results. This highly aligns with our motivation to ensure the temporal smoothness of the generated results. Inspired by BEAT, we leverage a pre-trained posture-based emotion classifier to evaluate the emotion transition effect in both datasets. As reported in Table~\ref{tab:EmoACC}, our method attains the best performance on emotion transition, which highly aligns with our visualization.

\begin{table}[h]
\centering
\caption{Ablation study on the adversarial loss in TED-ETrans dataset. w/o represents without adversarial loss in experiments.}
\vspace{-0.2em}
\label{tab:ablation_adversarial}
\renewcommand{\arraystretch}{0.9}
\footnotesize
\begin{tabular}{lcccc}
\toprule
Methods                         & $FGD_{h+t}$ $\downarrow$ & $FGD_{trans}$ $\downarrow$ & BC $\uparrow$   & Diversity $\uparrow$ \\ \midrule
Ours w/o                 & 15.31        & 32.72          & 0.802 & 79.64$^{\pm4.58}$ \\
\rowcolor[HTML]{ECF4FF}
\textbf{Ours}                     & \textbf{12.19}        & \textbf{23.54}          & \textbf{0.906} & \textbf{93.79$^{\pm2.53}$} \\ \bottomrule
\end{tabular}%
\end{table}

\begin{table}[h]
\centering

\caption{Comparison in emotion transition effect. EmoACC means whether the gestures in the head/tail represent the corresponding emotions.
}

\label{tab:EmoACC}
\footnotesize
\resizebox{\linewidth}{!}{
\begin{tabular}{lcccccccc}
\toprule
\multirow{2}{*}{Models} & \multicolumn{4}{c}{BEAT-ETrans}               & \multicolumn{4}{c}{TED-ETrans}               \\ \cmidrule(r){2-5} \cmidrule(r){6-9}
                        & FGD$_{h+t}$$\downarrow$ & FGD$_{trans}$$\downarrow$ & BC$\uparrow$   & EmoACC$\uparrow$    & FGD$_{h+t}$$\downarrow$  & FGD$_{trans}$$\downarrow$ & BC$\uparrow$   & EmoACC$\uparrow$  \\ \midrule 
Seq2Seq                 & 40.95 & 47.93          & 0.141 & 57.50  & 29.60 & 49.47          & 0.265 & 56.20 \\
S2G                  & 25.56 & 37.04          & 0.671 & 60.69  & 18.16 & 41.63          & 0.824 & 58.88 \\
Trimodal                & 14.09 & 42.50          & 0.764 & 69.81 & 21.06 & 33.20          & 0.758 & 63.82 \\
CAMN                & 9.03  & 27.53          & 0.794 & 72.87 & 19.28 & 41.04          & 0.785 & 74.55 \\
HA2G                   & 7.28  & 25.79          & 0.779 & 73.98 & 16.72 & 40.38          & 0.787 & 80.74 \\
DiffGesture            & 6.68  & 25.03          & 0.788 & 80.72 & 18.69 & 25.13          & 0.818 & 81.17 \\ \midrule
\rowcolor[HTML]{ECF4FF} 
\textbf{Ours} & \textbf{4.42} & \textbf{18.84} & \textbf{0.881} & \textbf{83.57} & \textbf{12.19} & \textbf{23.54} & \textbf{0.906} & \textbf{85.61} \\ 
\bottomrule
\end{tabular}
}

\end{table}

\subsection{Additional Visualization Results}
\label{sec:visualization}
\begin{figure*}[t]
\begin{center}
\includegraphics[width=0.98\linewidth]{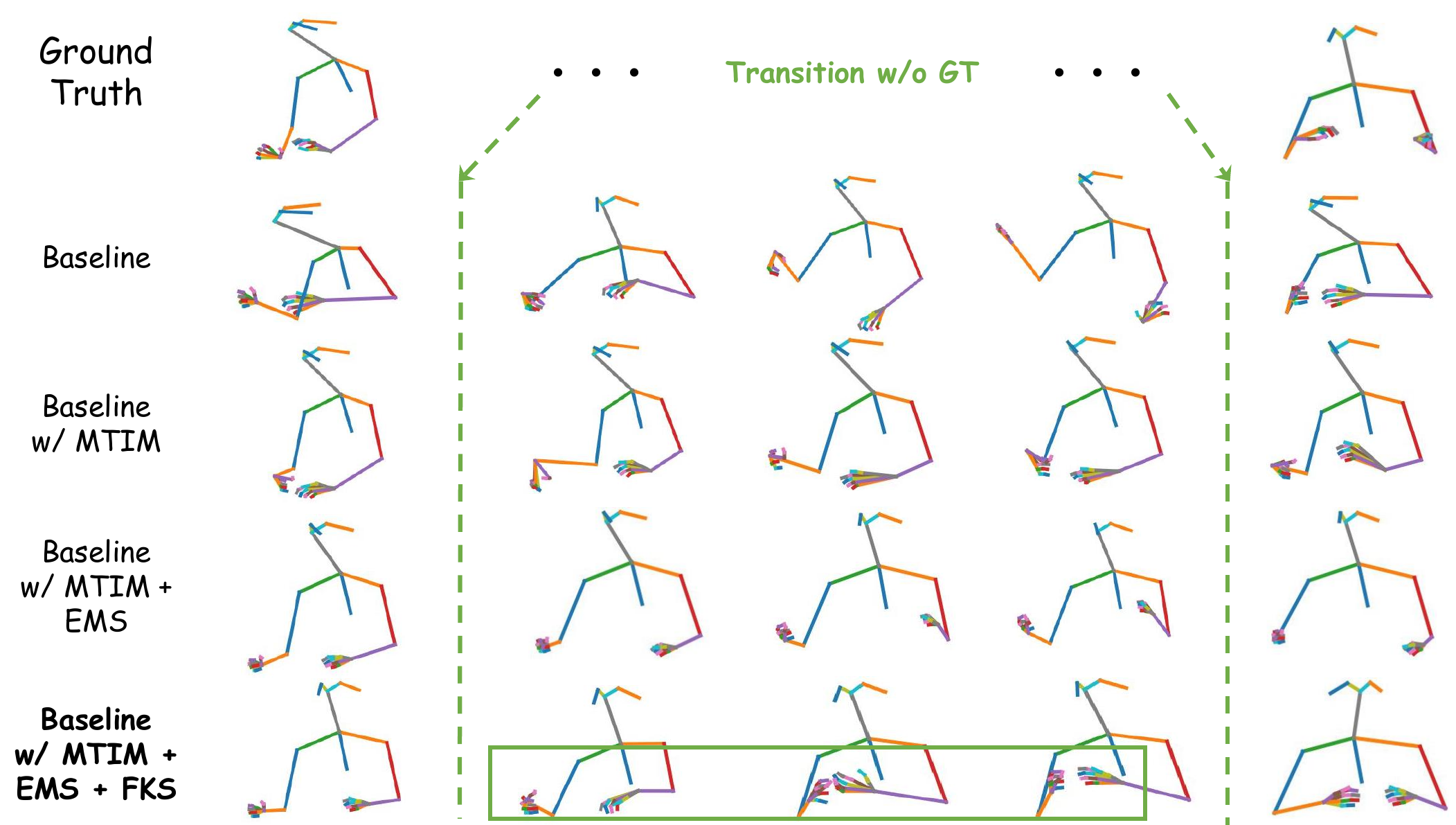}
\end{center}
\caption{Visual comparisons of ablation study on our newly collected \textbf{TED-ETrans dataset}. We show the key frames of the generated motions given the emotion transition of human speech. Best view on screen.
}
\label{fig:ablation2}
\end{figure*}

\begin{figure*}[t]
\begin{center}
\includegraphics[width=0.98\linewidth]{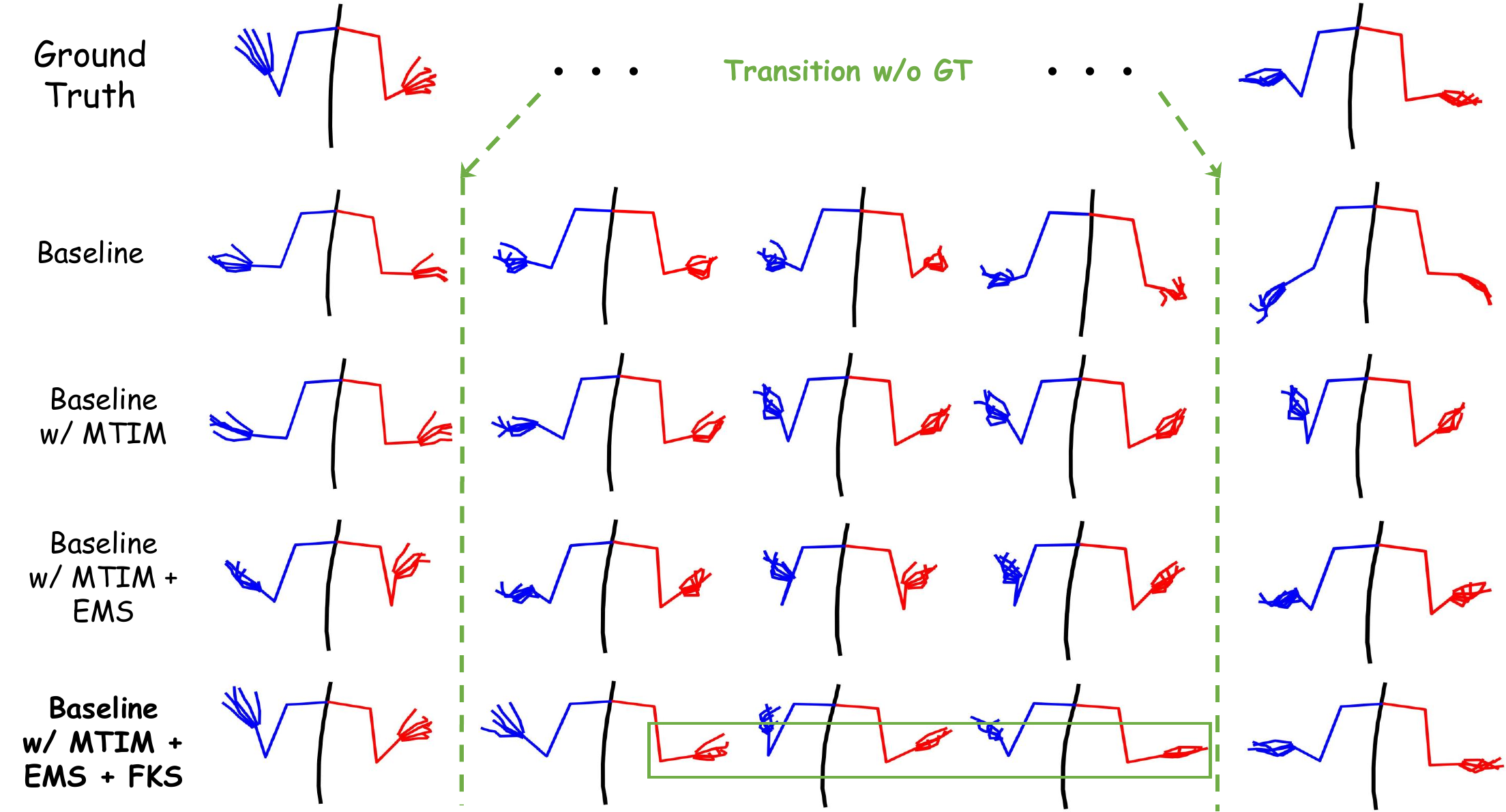}
\end{center}
\caption{Visual comparisons of ablation study on our newly collected \textbf{BEAT-ETrans dataset}. We show the key frames of the generated motions given the emotion transition of human speech. Best view on screen.
}
\label{fig:ablation1}
\end{figure*}

Here, we provide more visual results of our methods compared with other counterparts in the \emph{\textbf{demo video.}} Meanwhile, to fully demonstrate the effectiveness of our proposed components in the ablation study, we visualize vital frames of the synthesized gestures. As illustrated in Figure \ref{fig:ablation2} and Figure \ref{fig:ablation1}, we can clearly observe that all the combinations of our proposed components have positive impacts on the generated results.

\newpage
{
    \small
    \bibliographystyle{ieeenat_fullname}
    \bibliography{main}
}


\end{document}